%% file: main.tex
\documentclass[11pt]{article}

\usepackage[final]{acl}

\usepackage{times}
\usepackage{latexsym}
\usepackage[T1]{fontenc}
\usepackage[utf8]{inputenc}
\usepackage{microtype}
\usepackage{inconsolata}

\usepackage{amsmath}
\usepackage{amssymb}
\usepackage{amsthm}
\usepackage{booktabs}
\usepackage{multirow}
\usepackage{array}
\usepackage{xcolor}
\usepackage{comment}
\usepackage{url}
\usepackage{graphicx}

\graphicspath{{figs/}}

\usepackage{tikz}
\usetikzlibrary{positioning,arrows.meta,shapes.geometric,calc,fit,backgrounds}

\newcommand{\colfigwidth}{\columnwidth}

\newcommand{\LorE}{LorExperts}
\newcommand{\BTE}{BTExperts}

\title{Shape Mutating Expert Compression: LorExperts and BTExperts}

\author{
  \mdseries
  Inesh Chakrabarti\thanks{\ Equal contribution.} \quad
  Sourjya Roy\footnotemark[1] \quad
  Bowen Bao \\
  Thiago Crepaldi \quad
  Spandan Tiwari \quad
  Ashish Sirasao \\[0.6em]
  Advanced Micro Devices \\[0.3em]
  \texttt{\{inesh.chakrabarti,\,sourjya.roy,\,bowen.bao\}@amd.com} \\
  \texttt{\{thiago.crepaldi,\,spandan.tiwari,\,ashish.sirasao\}@amd.com}
}

\date{August 7, 2026}

\begin{document}
\maketitle

\input{sections/00_abstract}

\input{sections/01_introduction}
\input{sections/02_analysis}
\input{sections/03_lorexperts}
\input{sections/04_finetuning}
\input{sections/05_btexperts}
\input{sections/06_results}

\input{sections/07_discussion}
\input{sections/08_conclusion}

\bibliography{references/references}

\appendix
\input{appendices/C_ablations}
\input{appendices/B_coactivation}

\end{document}

%% file: sections/00_abstract.tex
\begin{abstract}
Mixture-of-Experts (MoE) language models deliver high capacity at low per-token
compute, but deploying them cheaply requires compressing their many expert
weight matrices. Expert \emph{pruning} (e.g., REAP) and \emph{merging} reduce
cost but sacrifice accuracy and require \emph{retraining the router}; low-rank
\emph{delta decomposition} of experts (e.g., D$^2$-MoE) preserves all experts and
the router, but degrades sharply as the expert count grows because a single
shared component cannot approximate many near-orthogonal experts.

Because MoE expert weights are near-orthogonal, a single shared component (as in
prior delta decomposition) scales poorly with the expert count; we show that
experts nonetheless organize into functional \emph{co-activation communities}
that are decoupled from weight similarity. Building on this, we introduce
\textbf{\LorE{}}, a \emph{router-preserving} compression method that clusters
experts, keeps one full-precision \emph{dominant} per cluster, and represents the
remaining members as low-rank corrections to their local dominant. \LorE{}
retains \emph{all} experts and the \emph{original router} (no router retraining).
At $\sim$50\% expert compression on Qwen3-30B-A3B and
Gemma-4-26B-A4B, \LorE{} preserves downstream accuracy and perplexity better than the baselines on most of the tasks; the margin over
D$^2$-MoE grows with expert count $E$.
We further give a reconstruction fine-tuning procedure for \LorE{}, and
\BTE{}, a tree organization of dominants and corrections that enables
inference-time amortization of shared computation.
\end{abstract}

%% file: sections/01_introduction.tex
\section{Introduction}
\label{sec:intro}

Large language models (LLMs) have improved rapidly as dense transformers have
scaled to hundreds of billions of parameters \citep{gpt3,palm,llama}. Scaling a
dense model, however, means paying for every parameter on every token. Sparse
Mixture-of-Experts (MoE) architectures ease this by adding capacity while holding
the cost of each forward pass roughly constant. An MoE layer keeps $E$ separate
experts, and for every token a small router activates only $k \ll E$ of them,
leaving the rest idle \citep{jiang2024mixtral,qwen3,deepseekmoe}. Because total capacity is decoupled
from the compute spent per token, models can grow to hundreds of billions of
parameters without a comparable rise in inference floating-point operations
(FLOPs).

The difficulty is that this capacity lives almost entirely in the experts, and
each expert is as large as a dense feed-forward block. Since only a few are
active per token, MoE inference is limited by the memory traffic of loading
expert weights rather than by arithmetic~\citep{moeoffload}. Recent architectures
push this further, favoring more experts that are each smaller and more
specialized, and configurations of $E{=}128$ are now
common~\citep{deepseekmoe,qwen3}. As this expert budget grows, serving the models
affordably comes down to compressing the experts. The aim is to shrink their
memory footprint without losing quality, and without disturbing the routing the
model has already learned.

\subsection{Related Work}
\label{sec:related}

\paragraph{Expert pruning.} Pruning removes low-importance experts. Candidates
are chosen by activation frequency or by an importance criterion, as in
REAP~\citep{reap}. The approach is simple and reduces both storage and per-token
compute. However, it \emph{discards model capacity} and \emph{changes the set of
experts the router can select}. The router is therefore miscalibrated after
pruning and must be adjusted or retrained. Accuracy also degrades on the
capabilities carried by the removed experts. The relative merit
of pruning and merging is benchmark-dependent. On discriminative metrics such as
perplexity and multiple-choice accuracy, merging has been reported to do
better~\citep{msmoe}. On generative tasks, REAP~\citep{reap} finds the opposite.
It attributes the gap to an irreducible error in merging, which arises from the
loss of independent routing control over the combined experts.

\paragraph{Expert merging.} Merging fuses similar experts into fewer experts.
M-SMoE~\citep{msmoe} groups experts by the cosine similarity of their router
logits and merges each group by frequency-weighted averaging. HC-SMoE~\citep{hcsmoe}
instead clusters experts by the similarity of their output activations, which
makes the grouping less dependent on routing statistics. Like pruning, merging
shrinks the expert set and usually requires the router to be re-calibrated or
retrained. Its quality depends on how losslessly similar experts can be fused.

\paragraph{Low-rank / delta decomposition.} A third family preserves \emph{all}
experts and the \emph{router}. It writes each expert as a shared component plus a
low-rank per-expert delta, in the spirit of LoRA~\citep{lora}.
D$^2$-MoE~\citep{d2moe} uses a shared (Fisher) mean
with per-expert singular value decomposition (SVD) deltas. SD-MoE~\citep{sdmoe} instead uses a spectral shared
component. Because the expert set and the routing are unchanged, \emph{no router
retraining is required}, which is attractive in practice. Their quality, though,
is bounded by how well a \emph{single} shared component plus a small delta can
approximate each expert.

\paragraph{Neuron permutation symmetry.} Feed-forward neurons have no canonical
ordering. Two functionally related networks can match closely only after a
permutation of their hidden units. This symmetry is well studied in model
merging, for example the activation and weight matching used in
Git~Re-Basin~\citep{gitrebasin}. We reuse this alignment machinery when forming
inter-expert residuals.

\subsection{Motivation}
\label{sec:motivation}

We build on delta decomposition, the most deployment-friendly of these families,
because it leaves every expert and the router in place. Its limitation is scale.
A \emph{single} shared component cannot represent a growing expert set, since one
global anchor sits far from most experts in a large, diverse pool. MoE expert
weights are also \emph{near-orthogonal}, so a small low-rank delta still leaves a
large residual. As a result, single-anchor methods degrade as $E$ grows
(Sec.~\ref{sec:analysis}).

We address this with two ideas. The first is to use \emph{multiple local
anchors}. Instead of one global root, we cluster the experts and keep a
full-precision \emph{dominant} per cluster, so every anchor stays close to its
members and the method scales with $E$. The second is to exploit permutation
symmetry. Two experts may be similar only \emph{up to a neuron permutation}, so
we \emph{align} each member's neurons to its dominant before forming the low-rank
residual (Sec.~\ref{sec:lorexperts}). Throughout, every expert and the
original router are left untouched, with no rerouting and no router retraining.

\paragraph{Contributions.}
\begin{itemize}
  \item \textbf{Analysis (Sec.~\ref{sec:analysis}).} We characterize the
        ``orthogonal-experts'' problem: why a single shared component scales
        poorly as $E$ grows, and we measure the functional structure that
        survives it---co-activation communities, and their projection-specific
        dissociation from weight similarity. We then quantify the two things a
        compression budget actually buys (Sec.~\ref{sec:budget}): expert spectra
        decay only slowly, so rank is expensive, while routing is strongly
        concentrated---the 64 busiest of 128 experts absorb $94.5\%$ of routed
        visits---so retaining a well-chosen minority exactly is cheap. Together
        these motivate the asymmetric budget \LorE{} uses.
  \item \textbf{\LorE{} (Sec.~\ref{sec:lorexperts}).} A router-preserving
        compression method that clusters experts, keeps one full-precision
        dominant per cluster, and encodes the remaining members as
        permutation-aligned low-rank corrections. Every expert and the original
        router are retained.
  \item \textbf{Fine-tuning (Sec.~\ref{sec:finetuning}).} A lightweight
        distillation step that fits only the low-rank factors to the original
        expert outputs, while the dominants and the router stay frozen.
  \item \textbf{\BTE{} (Sec.~\ref{sec:btexperts}).} A tree organization of
        dominants and corrections that amortizes shared computation at inference
        time.
\end{itemize}

%% file: sections/02_analysis.tex
\section{Analysis: The Orthogonal-Experts Problem}
\label{sec:analysis}

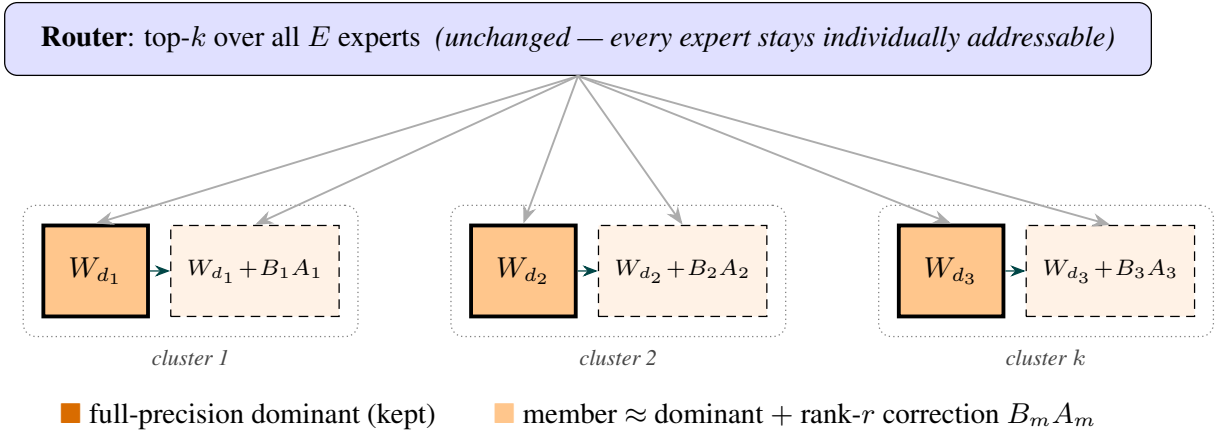
\begin{figure*}[t]
\centering
\resizebox{\textwidth}{!}{%
\begin{tikzpicture}[
  font=\small,>=Stealth,
  router/.style={draw,rounded corners,fill=blue!12,minimum width=12.6cm,minimum height=0.8cm,align=center},
  dom/.style={draw,very thick,fill=orange!45,minimum width=1.15cm,minimum height=1.0cm,align=center,font=\footnotesize},
  mem/.style={draw,densely dashed,fill=orange!10,minimum width=1.85cm,minimum height=1.0cm,align=center,font=\scriptsize},
  clab/.style={font=\scriptsize\itshape,text=gray!55!black},
  clbox/.style={draw,rounded corners,densely dotted,gray,inner sep=0.2cm},
  sel/.style={->,semithick,gray!65},
  shr/.style={->,densely dashed,teal!55!black}
]
\node[router] (R) at (-0.4,1.15)
  {\textbf{Router}: top-$k$ over all $E$ experts~ \emph{(unchanged --- every expert stays individually addressable)}};
\node[dom] (d1) at (-5.7,-1.4) {$W_{d_1}$};
\node[mem] (m1) at (-3.95,-1.4) {$W_{d_1}\!+\!B_1A_1$};
\node[clbox,fit=(d1)(m1)] (c1) {};
\node[clab,below=0pt of c1] {cluster 1};
\node[dom] (d2) at (-1.0,-1.4) {$W_{d_2}$};
\node[mem] (m2) at (0.75,-1.4) {$W_{d_2}\!+\!B_2A_2$};
\node[clbox,fit=(d2)(m2)] (c2) {};
\node[clab,below=0pt of c2] {cluster 2};
\node[dom] (d3) at (3.7,-1.4) {$W_{d_3}$};
\node[mem] (m3) at (5.45,-1.4) {$W_{d_3}\!+\!B_3A_3$};
\node[clbox,fit=(d3)(m3)] (c3) {};
\node[clab,below=0pt of c3] {cluster $k$};
\foreach \n in {d1,m1,d2,m2,d3,m3} \draw[sel] (R.south) -- (\n.north);
\draw[shr] (d1) -- (m1); \draw[shr] (d2) -- (m2); \draw[shr] (d3) -- (m3);
\node[font=\footnotesize] at (-0.4,-3.0)
  {\textcolor{orange!85!black}{$\blacksquare$}~full-precision dominant (kept)\qquad
   \textcolor{orange!45}{$\blacksquare$}~member $\approx$ dominant $+$ rank-$r$ correction $B_mA_m$};
\end{tikzpicture}%
}
\caption{\textbf{\LorE{} overview.} Experts are clustered (by co-activation or
weight similarity); each cluster keeps its highest-firing expert as a
full-precision \emph{dominant} $W_{d}$ and represents every other member as
$W_{d}+B_mA_m$ with a rank-$r$ correction. Unlike single-shared-component
decomposition, which uses one global root, \LorE{} uses $k$ \emph{local}
dominants, so approximation quality does not collapse as the expert count $E$
grows. The router and all $E$ experts are preserved---no rerouting or retraining.}
\label{fig:overview}
\end{figure*}

This section motivates \LorE{} (overview in Fig.~\ref{fig:overview}) by
characterizing \emph{why} shared-component low-rank decomposition of MoE experts
is hard, and identifying the structure that makes it tractable.

\paragraph{Setup: the single-anchor budget.} Let an MoE layer have $E$ experts
with weight matrices $W_e \in \mathbb{R}^{I \times H}$. A single-shared-component
method approximates $W_e \approx W_{\text{shared}} + \Delta_e$ with $\Delta_e$
low-rank, so its quality is bounded by how much of each expert the one shared
component can carry and how low-rank the leftover residual is. The difficulty is
an accounting one: a single anchor is amortized across the whole pool, so as $E$
grows it must sit close to ever more, ever more diverse experts at once. For
experts that are close to mutually orthogonal in weight space---as MoE experts
are, and as the co-activation analysis below takes as its starting point---any
one direction can align with only about $1/E$ of the pool's inter-expert
variance, leaving a near-full-rank residual for the per-expert delta to absorb.
Adding rank to $\Delta_e$ does not fix this, because the residual is not
low-rank to begin with. The remedy we pursue is therefore not a better global
anchor but \emph{more} anchors: $k$ local dominants, each close to its own
members (Sec.~\ref{sec:lorexperts}). What remains is to decide which experts
belong together.

\subsection{What the co-activation structure reveals}

\begin{figure*}[t]
\centering
\includegraphics[width=0.6\textwidth]{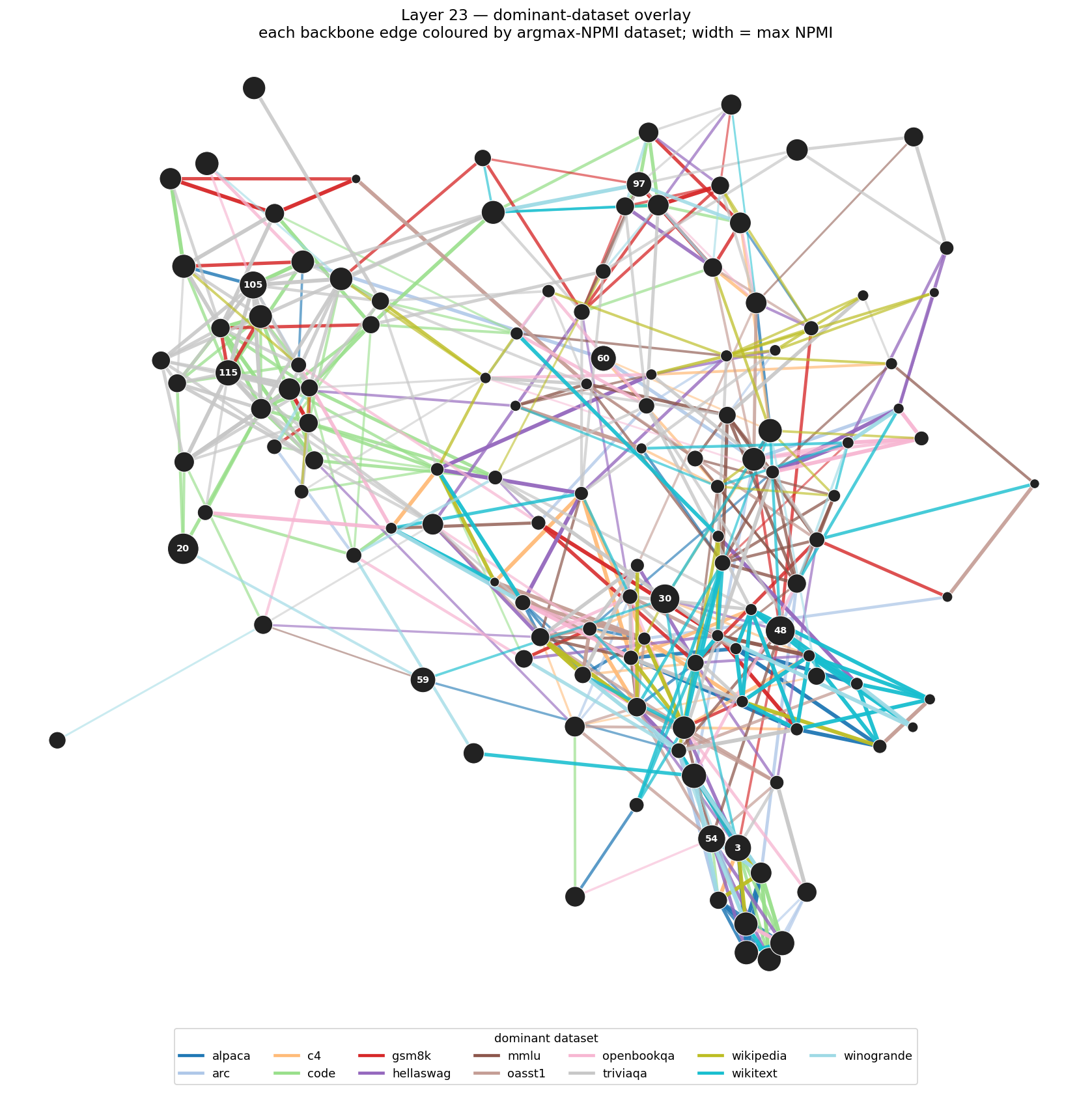}
\caption{Expert co-activation structure at Qwen3-30B-A3B layer~23 as a
  \emph{backbone graph}: experts are nodes, and each strong co-activation edge is
  colored by the dataset with the largest NPMI (code=green, math/GSM8K=red,
  web/wikitext=cyan, \dots) and widened by its strength. Experts partition into
  stable communities even though their weights are near-orthogonal, and distinct
  domains induce visibly specialized structure. A raw heatmap view and clique
  statistics appear in Appendix~\ref{sec:coact_appendix}.}
\label{fig:npmi}
\end{figure*}

Although expert \emph{weights} are near-orthogonal, the router induces rich,
non-random \emph{functional} structure. The expert co-activation graph (edges
weighted by normalized pointwise mutual information, NPMI), built across 13
datasets on Qwen3-30B-A3B (Fig.~\ref{fig:npmi}), exposes several
properties that bear directly
on the design choices---and failure modes---of prior compression methods.

\begin{itemize}
  \item \textbf{Stable communities.} Experts partition into consistent
        co-activation communities that persist across datasets---natural groups
        that \LorE{} clusters over (motivating the Coact-NPMI / $M_{\text{soft}}$
        metrics, Sec.~\ref{sec:lorexperts}).
  \item \textbf{Depth-varying granularity.} The number of co-activation
        communities grows with network depth. A \emph{uniform} compression budget
        is therefore suboptimal, motivating per-layer treatment.
  \item \textbf{Domain specialists.} Some communities are domain-general while
        others are strongly domain-specific: code and math (GSM8K) induce the
        most distinctive routing (visible as the colored communities in
        Fig.~\ref{fig:npmi}), whereas web text (C4, Wikipedia) tracks the
        consensus. These specialists are \emph{rare} yet load-bearing for their
        domain.
\end{itemize}

\paragraph{Consequences for prior methods---and for \LorE{}.} This structure is
exactly what the prior families endanger. Frequency-based \emph{pruning}
(Sec.~\ref{sec:related}) discards precisely the rare domain specialists that
carry code/math ability, and \emph{merging} (Sec.~\ref{sec:related}) blurs
distinct communities---both then force router re-calibration. \LorE{} instead
keeps every expert and the router intact, preserving this functional structure
by construction. Moreover, the structure lives in the \emph{routing} while the
\emph{weights} are near-orthogonal---a decoupling we measure directly
(Sec.~\ref{sec:dissociation}, Table~\ref{tab:dissociation})---so
single-shared-component weight decomposition (Sec.~\ref{sec:related}) cannot
exploit it, which is why \LorE{} uses local per-cluster dominants rather than one
global root.

\subsection{The functional--weight dissociation is projection-specific}
\label{sec:dissociation}

Do co-firing experts also have similar \emph{weights}? We correlate each expert
pair's co-activation (NPMI) with its weight cosine similarity, per layer and per
projection (Table~\ref{tab:dissociation}). The correlation is small but
consistently significant for the \texttt{gate} projection
($r\approx0.08$--$0.21$, $p<10^{-3}$) and effectively zero for \texttt{up} and
\texttt{down}.

\begin{table}[t]
\centering
\caption{Correlation between expert co-activation (NPMI) and weight cosine
  similarity, by layer and projection (Qwen3-30B-A3B). $^{**}p<10^{-3}$,
  $^{*}p<0.05$.}
\label{tab:dissociation}
\small
\begin{tabular}{lrrrr}
\toprule
Layer & \texttt{gate} & \texttt{up} & \texttt{down} & mean \\
\midrule
L5  & $+0.208^{**}$ & $+0.017$ & $-0.026^{*}$ & $+0.164^{**}$ \\
L11 & $+0.085^{**}$ & $+0.010$ & $+0.006$      & $+0.078^{**}$ \\
L23 & $+0.133^{**}$ & $-0.005$ & $+0.009$      & $+0.115^{**}$ \\
L35 & $+0.178^{**}$ & $-0.001$ & $+0.020$      & $+0.160^{**}$ \\
L47 & $+0.195^{**}$ & $+0.007$ & $+0.011$      & $+0.167^{**}$ \\
\bottomrule
\end{tabular}
\end{table}

The pattern is interpretable. The \texttt{gate} projection selects \emph{which
inputs activate} an expert, so co-firing experts share a \emph{weak} alignment in
what excites them; their actual \emph{computation} (\texttt{up}, \texttt{down}) is
essentially orthogonal. Experts that fire together thus respond to similar
contexts but compute different things---\emph{complementary, not redundant}. Even
the gate correlation is small ($r\le0.21$), so experts remain largely
near-orthogonal in weight space. This is precisely why routing-space structure is
not recoverable by a shared-component weight decomposition, and why \LorE{}
corrects each expert against a \emph{local} dominant rather than assuming a single
shared low-rank basis across experts.

\subsection{What the budget can buy: spectra and routing mass}
\label{sec:budget}

Two measurable quantities decide how a compression budget for an MoE layer is
best spent: how fast expert spectra decay, which sets what a unit of rank buys,
and how concentrated the routing is, which sets what keeping a single expert
exact buys. Measured on Qwen3-30B-A3B, together they argue for exactly the
asymmetric allocation \LorE{} adopts.

\paragraph{Spectra decay, but slowly.} Table~\ref{tab:spectra} reports the
fraction of squared Frobenius energy a rank-$r$ truncation retains, averaged over
the three projections of 12 experts at layers 5, 23 and 40. There \emph{is}
genuine low-rank structure: rank 64 retains $26.0\%$ where a perfectly flat
spectrum would retain only $64/768 = 8.3\%$. But the decay is shallow---reaching
$72\%$ of the energy already costs rank 311, i.e.\ $40\%$ of the maximum rank
768. Rank is expensive on these matrices, so a budget spread thinly and uniformly
across all $E$ experts buys comparatively little per expert; it is worth asking
whether some of it is better spent keeping selected experts exactly.

\begin{table}[t]
\centering
\caption{Fraction of squared Frobenius energy retained by a rank-$r$ truncation
  of an expert matrix ($I{=}768$, $H{=}2048$, so rank $768$ is lossless), mean
  over \texttt{gate}/\texttt{up}/\texttt{down} of 12 experts per layer. The last
  row is what a flat spectrum ($r/768$) would give.}
\label{tab:spectra}
\small
\setlength{\tabcolsep}{3pt}
\begin{tabular}{lrrrrr}
\toprule
Layer & $r{=}64$ & $r{=}200$ & $r{=}311$ & $r{=}450$ & $r{=}558$ \\
\midrule
L5           & $0.266$ & $0.565$ & $0.724$ & $0.860$ & $0.929$ \\
L23          & $0.266$ & $0.566$ & $0.725$ & $0.860$ & $0.929$ \\
L40          & $0.248$ & $0.548$ & $0.712$ & $0.853$ & $0.926$ \\
\midrule
mean         & $0.260$ & $0.560$ & $0.720$ & $0.858$ & $0.928$ \\
flat spectrum & $0.083$ & $0.260$ & $0.405$ & $0.586$ & $0.727$ \\
\bottomrule
\end{tabular}
\end{table}

\paragraph{Routing is strongly concentrated.} Expert utilization is meanwhile far
from uniform. Over five layers, the 64 busiest of the 128 experts absorb $94.5\%$
of routed token--expert visits, against the $50\%$ that uniform routing would
give, and 18--30 experts per layer are effectively dead ($<10^{-4}$ of traffic).
Concentration this strong is what makes an asymmetric budget attractive: if a
minority of experts carries almost all of the traffic, then holding exactly those
experts in full precision costs a well-understood share of the budget and removes
approximation error where it is encountered most often.

\begin{table}[t]
\centering
\caption{Routed token--expert visits absorbed by the $64$ busiest of $E{=}128$
  experts, measured on WikiText-2. Uniform routing would give $0.500$.
  ``Dead'' counts experts receiving $<10^{-4}$ of traffic.}
\label{tab:coverage}
\small
\begin{tabular}{lrr}
\toprule
Layer & top-64 share & dead experts \\
\midrule
L5   & $0.928$ & $18$ \\
L12  & $0.950$ & $22$ \\
L23  & $0.942$ & $30$ \\
L31  & $0.961$ & $27$ \\
L40  & $0.944$ & $26$ \\
\midrule
mean & $0.945$ & $24.6$ \\
\bottomrule
\end{tabular}
\end{table}

Taken together the two measurements motivate the shape of \LorE{}: because rank
is expensive, uniform low-rank treatment of every expert is a poor use of the
budget; because routing is concentrated, keeping a well-chosen minority exact
covers most of what the router actually does. Retaining a full-precision dominant
per cluster and spending only low-rank corrections on the remainder follows
directly, and it is why \LorE{} selects each cluster's highest-firing member
rather than its geometric center.

One caveat qualifies how the dominant set should be read. Utilization here is
measured on WikiText-2, and Sec.~\ref{sec:analysis} shows that code and math
induce visibly different communities, so the busy set is partly domain-dependent.
This is a reason to keep every expert addressable rather than pruning the tail:
the rare domain specialists are precisely the experts a single-domain utilization
estimate will rank last.

\subsection{Implication for method design}

Two consequences drive \LorE{}. First, use \emph{multiple local anchors}
(per-cluster dominants) rather than one global root, so each anchor is close to
its members and the residual it must correct is small---and the approach scales
with $E$ instead of degrading (ablated in Appendix~\ref{sec:ablations}).
Second, by Sec.~\ref{sec:budget}, split the budget asymmetrically rather than
uniformly: rank buys little on these spectra, while routing mass is concentrated
enough that full-precision dominants pay for themselves---so a dominant should be
chosen for the traffic it absorbs.

%% file: sections/03_lorexperts.tex
\section{\LorE{}: Router-Preserving Low-Rank Expert Compression}
\label{sec:lorexperts}

\subsection{Representation}

Consider one MoE layer with $E$ experts, each a set of weight matrices (gate, up,
down) that we write jointly as $W_e$. \LorE{} partitions the experts into $k$
clusters. Within each cluster it keeps one \emph{dominant} expert at full
precision and represents every other member $m$ as a low-rank correction of its
dominant,
\[
  \begin{gathered}
    W_m \;\approx\; W_{\text{dominant}} + B_m A_m , \\
    \operatorname{rank}(B_m A_m) = r ,
  \end{gathered}
\]
with factors $B_m \in \mathbb{R}^{d \times r}$, $A_m \in \mathbb{R}^{r \times d'}$.
The dominant is stored exactly; each member costs only $B_m, A_m$. With
$k \approx E/2$ and $r=64$, expert parameters are reduced by roughly one half; the
ratio is controlled by $k$ and $r$.

\subsection{Compression pipeline}

\paragraph{Stage 1 --- Router profiling.} We pass a small calibration set (C4, 64
sequences of length 2048) through the model with router hooks, accumulating
per-layer firing counts $N_e$. These counts seed clustering and dominant
selection. Profiling only reads the router: \LorE{} keeps all $E$ experts and leaves the
gate unchanged, unlike pruning or merging.

\paragraph{Stage 2 --- Clustering.} Per layer, we form a pairwise distance matrix
$D = 1 - M$ with the diagonal zeroed. We then run frequency-seeded $k$-medoids:
the medoids start at the highest-firing experts, and the assign/update loop
iterates to convergence. Seeding from frequently-used experts biases the
dominants toward experts the router actually selects. Each cluster's dominant is
its highest-firing member, with ties broken by co-activation centrality, and the
remaining members are queued for reconstruction. An optional \emph{protection
set} can be pulled out as singleton clusters and kept at full precision. We
compare three choices of the distance $M$ empirically (Sec.~\ref{sec:results}):
weight cosine (WS-Frob), co-activation NPMI (Coact-NPMI), and soft co-activation
($M_{\text{soft}}$).

\paragraph{Stage 3 --- Alignment and low-rank decomposition.} We apply three
steps to each non-dominant member.
\emph{(i) Neuron alignment.} An expert's hidden feed-forward network (FFN) has no
canonical ordering of its intermediate neurons, so a member may store
corresponding neurons at different indices than its dominant. This inflates the naive residual
$W_m - W_{\text{dominant}}$, which can be full-rank even when the experts match up
to a permutation. Aligning networks by such permutations is well established in
model merging~\citep{gitrebasin}, and has been applied to experts before merging
in MoE compression~\citep{msmoe}. We therefore align the member to its dominant
before decomposing. Treating each neuron as the concatenation of its gate, up, and down
columns, we score neuron $i$ of the dominant against neuron $j$ of the member by
the squared distance
\begin{equation}
  C_{ij} = \bigl\lVert a_i - b_j \bigr\rVert_2^2
         = \lVert a_i \rVert^2 + \lVert b_j \rVert^2 - 2\, a_i^\top b_j ,
  \label{eq:cost}
\end{equation}
where $a_i = W_{\text{dom}}[:,i]$ and $b_j = W_m[:,j]$ are the concatenated
neuron vectors. We then seek the permutation $P$ of the member's neurons that
minimizes the total matching cost,
\begin{equation}
  P^\star = \arg\min_{P} \sum_i C_{i,\,P(i)} ,
  \label{eq:assign}
\end{equation}
which we solve exactly with the Jonker--Volgenant algorithm. Applying $P^\star$
gives the aligned member $W_m^{P}$. The reported results use this alignment.

\emph{(ii) Truncated-SVD residual.} We form the aligned residual and take its
rank-$r$ SVD,
\begin{equation}
  R = W_m^{P} - W_{\text{dominant}} \approx U_r \Sigma_r V_r^\top ,
  \label{eq:svd}
\end{equation}
computed with a randomized solver ($r=64$). Splitting the singular values
symmetrically yields the low-rank factors
\begin{equation}
  \begin{gathered}
    B_m = U_r \Sigma_r^{1/2} , \qquad A_m = \Sigma_r^{1/2} V_r^\top , \\
    \text{so } B_m A_m \approx R .
  \end{gathered}
  \label{eq:factors}
\end{equation}

\emph{(iii) Installation.} The member is replaced by a module that computes
$W_{\text{dominant}} + B_m A_m$ at inference, while the dominants and the router
remain untouched.

\paragraph{Stage 4 --- Reconstruction fine-tuning.} A short output-matching pass
refines the factors (Sec.~\ref{sec:finetuning}); it updates only $B_m, A_m$.

\paragraph{Variants.} The framework is parameterized by the clustering distance
$M$ and an optional protection set, giving four variants we evaluate side by side
(Sec.~\ref{sec:results}, Tables~\ref{tab:qwen3}--\ref{tab:gemma4}):
\begin{itemize}\itemsep2pt
  \item \textbf{WS-Frob} clusters on the weight-space Frobenius distance
    $\lVert W_i-W_j\rVert_F$. It uses no router information, so it serves as a
    routing-agnostic reference.
  \item \textbf{WS-Frob+protK8} uses the same weight-space clustering but keeps
    eight full-precision protected experts per layer. This isolates the effect of
    protection on an otherwise routing-agnostic variant.
  \item \textbf{Coact-NPMI} clusters on single-mixture routing co-activation,
    measured as $1-\mathrm{NPMI}$ over a mixed calibration set. It is our first
    routing-aware variant.
  \item \textbf{$M_{\text{soft}}$} clusters on the cross-domain soft consensus
    $1-M_{\text{soft}}$, optionally with the router-identified protection list,
    and is our strongest variant. Because $k$ and the protection set are tunable,
    we report it at several operating points to trace the compression--quality
    frontier.
\end{itemize}
All variants share the pipeline above and differ only in the Stage-2 distance and
whether a protection set is supplied.

\subsection{Storage}

For a cluster with a dominant and $c$ members, storage is one full matrix plus
$c$ rank-$r$ factor pairs, versus $c{+}1$ full matrices uncompressed. Summed over
clusters, the layer's expert parameters scale with the number of dominants (set
by $k$) plus $r$ times the number of members. Thus $k$ and $r$ jointly set the
compression ratio, and every expert remains individually addressable by the
unchanged router.

%% file: sections/04_finetuning.tex
\section{Reconstruction Fine-Tuning for \LorE{}}
\label{sec:finetuning}

Truncated-SVD (Stage 3) minimizes weight-space error; a short output-matching
refinement then recovers the error that matters for the token distribution---at
negligible cost.

\paragraph{Objective.} On \emph{cached} calibration activations $x$, we fit each
member's factors so its output matches the original expert's,
\[
  \min_{B_m, A_m}\;
  \big\lVert (W_{\text{dominant}} + B_m A_m)\,x - W_m x \big\rVert_F^2 ,
\]
training \emph{only} the rank-$r$ factors $B_m, A_m$; the dominant weights and the
router are frozen.

\paragraph{Why it is cheap.} Three properties make this the least costly stage of
\LorE{}:
\emph{(i)} it updates only the small rank-$r$ factors---not the base model and not
the router, so no gradients flow through either;
\emph{(ii)} the objective is \emph{local and per-member} and is evaluated on
\emph{pre-cached} activations, so there are no full-model forward/backward passes
during fitting---just small matmuls;
\emph{(iii)} it is an \emph{activation-weighted low-rank regression}, which admits
a near-closed-form solution: whitening by the activation second moment and taking
a (generalized) SVD of the residual recovers the optimal rank-$r$ factors
directly, in place of iterative gradient steps.

Its wall-clock cost, and its quality contribution relative to the SVD
initialization (Stage-3-only vs.\ Stage-3$+$4), are reported with the calibration
cost (Sec.~\ref{sec:results}). We expect the SVD initialization to do the heavy
lifting and this stage to be a cheap refinement.

%% file: sections/05_btexperts.tex
\section{\BTE{}: Tree Organization for Inference-Time Amortization}
\label{sec:btexperts}

\BTE{} organizes the dominants and their corrections in a binary tree. We
emphasize up front what \BTE{} is \emph{not}: it does \emph{not} improve
compression \emph{quality}. On near-orthogonal experts, sharing corrections
across a hierarchy does not beat flat per-cluster decomposition: in a controlled
study on Mixtral-8$\times$7B ($E{=}8$) at matched $7.5\times$ compression, the
hierarchical organization matches flat decomposition within noise in perplexity
(PPL; $\Delta\text{PPL}=0.099$ vs.\ $0.098$ after reconstruction fine-tuning; $0.112$
vs.\ $0.105$ zero-shot, 3 seeds; Appendix~\ref{sec:ablations}). Its sole purpose
is therefore \emph{inference-time amortization}: components shared by several
selected experts are computed once and reused.

\subsection{Amortized forward pass}

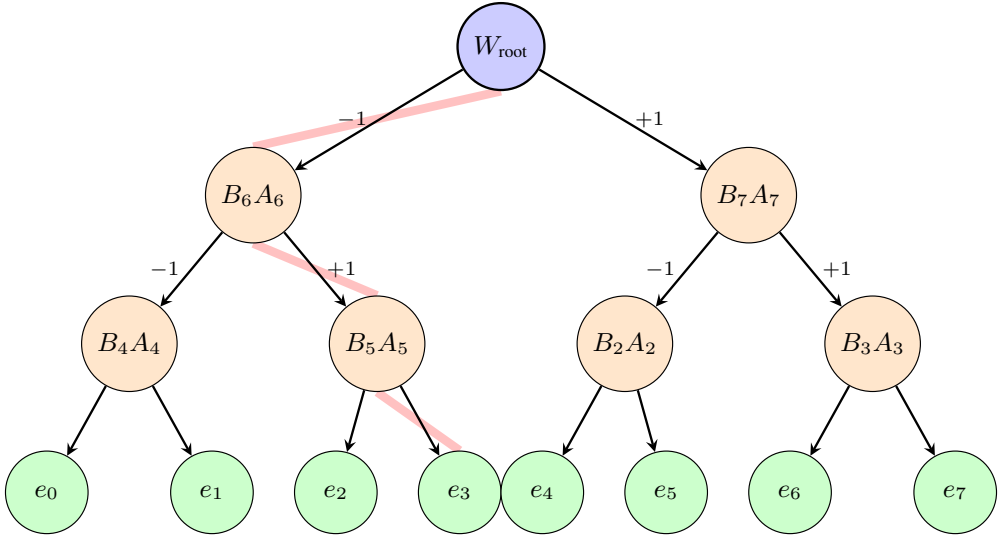
\begin{figure*}[t]
\centering
\resizebox{0.82\textwidth}{!}{%
\begin{tikzpicture}[
  node/.style={circle,draw,minimum size=1.0cm,font=\small},
  root/.style={node,fill=blue!20,thick},
  internal/.style={node,fill=orange!20},
  leaf/.style={node,fill=green!20},
  edge/.style={->,>=stealth,thick},
  lbl/.style={font=\scriptsize,midway}
]
\node[root] (R) at (0,0) {$W_{\text{root}}$};
\node[internal] (L1) at (-3,-1.8) {$B_6A_6$};
\node[internal] (L2) at ( 3,-1.8) {$B_7A_7$};
\draw[edge] (R) -- node[lbl,left]{$-1$} (L1);
\draw[edge] (R) -- node[lbl,right]{$+1$} (L2);
\node[internal] (LL) at (-4.5,-3.6) {$B_4A_4$};
\node[internal] (LR) at (-1.5,-3.6) {$B_5A_5$};
\node[internal] (RL) at ( 1.5,-3.6) {$B_2A_2$};
\node[internal] (RR) at ( 4.5,-3.6) {$B_3A_3$};
\draw[edge] (L1) -- node[lbl,left]{$-1$} (LL);
\draw[edge] (L1) -- node[lbl,right]{$+1$} (LR);
\draw[edge] (L2) -- node[lbl,left]{$-1$} (RL);
\draw[edge] (L2) -- node[lbl,right]{$+1$} (RR);
\node[leaf] (e0) at (-5.5,-5.4) {$e_0$};
\node[leaf] (e1) at (-3.5,-5.4) {$e_1$};
\node[leaf] (e2) at (-2.0,-5.4) {$e_2$};
\node[leaf] (e3) at (-0.5,-5.4) {$e_3$};
\node[leaf] (e4) at ( 0.5,-5.4) {$e_4$};
\node[leaf] (e5) at ( 2.0,-5.4) {$e_5$};
\node[leaf] (e6) at ( 3.5,-5.4) {$e_6$};
\node[leaf] (e7) at ( 5.5,-5.4) {$e_7$};
\draw[edge] (LL) -- (e0); \draw[edge] (LL) -- (e1);
\draw[edge] (LR) -- (e2); \draw[edge] (LR) -- (e3);
\draw[edge] (RL) -- (e4); \draw[edge] (RL) -- (e5);
\draw[edge] (RR) -- (e6); \draw[edge] (RR) -- (e7);
\begin{scope}[on background layer]
  \draw[red!60,line width=3pt,opacity=0.4]
    (R.south) -- (L1.north) (L1.south) -- (LR.north) (LR.south) -- (e3.north);
\end{scope}
\end{tikzpicture}%
}
\caption{\BTE{} tree ($E{=}8$). Shared components near the root are computed once
  and amortized across experts that share a path; leaves are experts. The
  highlighted red path is path($e_3$). The tree is an \emph{inference-time}
  organization, not a compression-quality mechanism.}
\label{fig:tree}
\end{figure*}

When multiple selected experts share a cluster (dominant) or a tree ancestor,
the shared computation---$x W_{\text{dominant}}^\top$ and any shared correction
factors---is computed \emph{once} per token and broadcast to those experts,
rather than recomputed per expert. The tree (Fig.~\ref{fig:tree}) makes this
sharing explicit: experts on the same path share the ancestors' contributions.

\subsection{Inference cost: theoretical FLOPs}

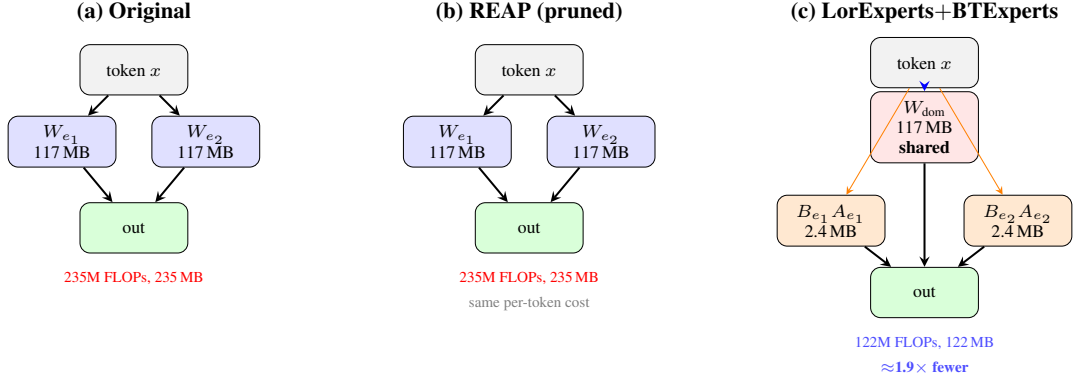
\begin{figure*}[t]
\centering
\resizebox{0.88\textwidth}{!}{%
\begin{tikzpicture}[
  box/.style={draw,rounded corners,minimum width=1.5cm,minimum height=0.7cm,font=\scriptsize,align=center},
  sbox/.style={box,fill=blue!12},
  lbox/.style={box,fill=orange!18},
  rbox/.style={box,fill=red!10},
  arr/.style={->,>=stealth,thick},
  tarr/.style={->,>=stealth},
  lab/.style={font=\tiny}
]
\node[font=\small\bfseries] at (-5.5,1.5) {(a) Original};
\node[box,fill=gray!10] (t1) at (-5.5,0.7) {token $x$};
\node[sbox] (a1) at (-6.5,-0.3) {$W_{e_1}$\\117\,MB};
\node[sbox] (a2) at (-4.5,-0.3) {$W_{e_2}$\\117\,MB};
\draw[arr] (t1)--(a1); \draw[arr] (t1)--(a2);
\node[box,fill=green!15] (o1) at (-5.5,-1.5) {out};
\draw[arr] (a1)--(o1); \draw[arr] (a2)--(o1);
\node[lab,red] at (-5.5,-2.2) {235M FLOPs, 235\,MB};
\node[font=\small\bfseries] at (0,1.5) {(b) REAP (pruned)};
\node[box,fill=gray!10] (t2) at (0,0.7) {token $x$};
\node[sbox] (b1) at (-1.0,-0.3) {$W_{e_1}$\\117\,MB};
\node[sbox] (b2) at (1.0,-0.3) {$W_{e_2}$\\117\,MB};
\draw[arr] (t2)--(b1); \draw[arr] (t2)--(b2);
\node[box,fill=green!15] (o2) at (0,-1.5) {out};
\draw[arr] (b1)--(o2); \draw[arr] (b2)--(o2);
\node[lab,red] at (0,-2.2) {235M FLOPs, 235\,MB};
\node[lab,gray] at (0,-2.55) {same per-token cost};
\node[font=\small\bfseries] at (5.5,1.5) {(c) \LorE{}$+$\BTE{}};
\node[box,fill=gray!10] (t3) at (5.5,0.8) {token $x$};
\node[rbox] (r3) at (5.5,-0.1) {$W_{\text{dom}}$\\117\,MB\\\textbf{shared}};
\draw[arr,blue] (t3)--(r3);
\node[lbox] (c1) at (4.2,-1.4) {$B_{e_1}A_{e_1}$\\2.4\,MB};
\node[lbox] (c2) at (6.8,-1.4) {$B_{e_2}A_{e_2}$\\2.4\,MB};
\draw[tarr,orange] (t3)--(c1); \draw[tarr,orange] (t3)--(c2);
\node[box,fill=green!15] (o3) at (5.5,-2.4) {out};
\draw[arr] (r3)--(o3); \draw[arr] (c1)--(o3); \draw[arr] (c2)--(o3);
\node[lab,blue!70] at (5.5,-3.1) {122M FLOPs, 122\,MB};
\node[lab,blue!70] at (5.5,-3.45) {\textbf{$\approx$1.9$\times$ fewer}};
\end{tikzpicture}%
}
\caption{Per-token forward pass (gate\_proj, top-2 routing). \textbf{(a)} Original:
two full expert matmuls. \textbf{(b)} REAP: identical per-token cost---pruning
removes experts but top-$k$ still runs $k$ full survivors. \textbf{(c)}
\LorE{}$+$\BTE{} for two co-selected experts sharing a dominant: the dominant is
computed once and reused, plus two cheap rank-$r$ corrections, cutting FLOPs and
HBM reads $\approx 1.9\times$ (kernel-independent). Adapted from the prior draft's
FLOP analysis.}
\label{fig:flops}
\end{figure*}

Standard top-$k$ routing evaluates $k$ full experts per token, costing
$k\cdot 2IH$ FLOPs and loading $k$ full weight tensors. \emph{Pruning} (REAP)
does not reduce this: top-$k$ still selects $k$ full \emph{survivors}, so
per-token FLOPs and high-bandwidth memory (HBM) reads are unchanged (a $0\times$ reduction---pruning saves
storage, not per-token compute). \BTE{} instead \emph{amortizes}: when co-selected
experts share a dominant, that dominant is computed \emph{once} and only the cheap
rank-$r$ corrections are added per expert (Fig.~\ref{fig:flops}). For such a group,
\[
  \text{speedup} \;=\; \frac{k\cdot 2IH}{2IH + k\cdot 2r(I+H)} \;\approx\; 1.9\times
\]
for Mixtral-8$\times$7B gate\_proj ($k{=}2$, $I{=}14336$, $H{=}4096$, $r{=}64$).
Because this projection is memory-bandwidth-bound (arithmetic intensity
$\approx 1$ FLOP/byte), the reduction is \emph{kernel-independent}. Two caveats
keep it honest: the benefit is \emph{routing-dependent} (it applies only when the
router co-selects experts from the same cluster), and it is \emph{shared with any
single-shared-component method}---D$^2$-MoE can amortize its shared base likewise.\looseness=1

%% file: sections/06_results.tex
\section{Results}
\label{sec:results}

We evaluate on Qwen3-30B-A3B ($E{=}128$) and Gemma-4-26B-A4B at $\sim$50\% expert
compression, reporting perplexity on Wikitext-103 and downstream accuracy via the
LM Evaluation Harness (MMLU, ARC-C, HellaSwag, WinoGrande, PIQA). We compare
against merging (M-SMoE, HC-SMoE), pruning (Freq-Prune, REAP), and delta
decomposition (D$^2$-MoE), all at matched compression
(Tables~\ref{tab:qwen3}--\ref{tab:gemma4}).

\begin{table*}[t]
  \centering
  \caption{Qwen3-30B-A3B ($\sim$50\% expert compression; uncompressed PPL $=8.50$).}
  \label{tab:qwen3}
  \small
  \begin{tabular}{l l r r r r r r}
  \toprule
  Method & Family & PPL $\downarrow$ & MMLU & ARC-C & HellaS & WinoG & PIQA \\
  \midrule
  Uncompressed & --- & 8.50 & 77.8 & 52.6 & 59.6 & 71.0 & 79.4 \\
  \midrule
  M-SMoE       & merge & 68.61 & 24.7 & 20.8 & 32.1 & 55.1 & 60.4 \\
  HC-SMoE      & merge & 39.50 & 43.3 & 27.6 & 32.9 & 54.1 & 61.4 \\
  Freq-Prune   & prune & 29.64 & 44.6 & 21.9 & 52.9 & 63.2 & 65.1 \\
  REAP         & prune & 34.14 & 32.1 & 23.5 & 38.9 & 54.5 & 63.3 \\
  D\textsuperscript{2}-MoE & delta & 23.58 & 45.7 & 40.8 & 51.7 & 66.9 & 70.5 \\
  \midrule
  \LorE{} (WS-Frob)          & ours & 12.36 & 42.4 & 32.4 & 55.5 & 71.1 & 76.4 \\
  \LorE{} (WS-Frob+protK8)   & ours & 11.79 & 53.7 & 41.1 & 55.7 & 71.0 & 76.9 \\
  \LorE{} (Coact-NPMI)       & ours & 12.20 & 54.1 & 37.1 & 52.6 & 68.4 & 75.3 \\
  \LorE{} (M\textsubscript{soft}) @40.6\% & ours & 11.99 & 54.9 & 39.1 & 52.4 & 68.8 & 74.3 \\
  \LorE{} (M\textsubscript{soft}) @51\%   & ours & 15.37 & 46.0 & 28.8 & 45.5 & 63.8 & 68.4 \\
  \bottomrule
  \end{tabular}
\end{table*}

\begin{table*}[t]
  \centering
  \caption{Gemma-4-26B-A4B ($\sim$50\% expert compression; uncompressed PPL $\approx 7.15$).}
  \label{tab:gemma4}
  \small
  \begin{tabular}{l l r r r r r r}
  \toprule
  Method & Family & PPL $\downarrow$ & MMLU & ARC-C & HellaS & WinoG & PIQA \\
  \midrule
  Uncompressed & --- & $\sim$7.15 & 74.3 & 64.5 & 63.4 & 76.1 & 82.2 \\
  \midrule
  M-SMoE @50\%       & merge & 14.94 & 27.8 & 38.9 & 74.8 & 73.7 & 79.2 \\
  HC-SMoE @50\%      & merge & 25.20 & 35.6 & 30.0 & 46.4 & 55.2 & 64.1 \\
  Freq-Prune @50\%   & prune & 11.57 & 34.8 & 48.5 & 79.6 & 74.6 & 81.3 \\
  REAP @50\%         & prune & 15.09 & 27.4 & 36.3 & 64.9 & 63.5 & 72.2 \\
  D\textsuperscript{2}-MoE @50\% & delta & 21.59 & 39.0 & 37.4 & 56.4 & 66.7 & 70.3 \\
  \midrule
  \LorE{} (WS-Frob) @49.9\%        & ours & 13.50 & 26.3 & 36.4 & 74.4 & 74.4 & 78.0 \\
  \LorE{} (WS-Frob+protK8) @49.9\% & ours & 14.66 & 51.1 & 45.2 & 71.9 & 72.0 & 76.0 \\
  \LorE{} (Coact-NPMI) @49.9\%     & ours & 14.42 & 50.1 & 42.7 & 55.5 & 68.0 & 68.8 \\
  \LorE{} (M\textsubscript{soft}) @41.4\% & ours & 13.32 & 54.6 & 50.1 & 63.9 & 70.6 & 72.9 \\
  \LorE{} (M\textsubscript{soft}) @51.6\% & ours & 18.54 & 47.4 & 38.8 & 48.4 & 62.4 & 65.8 \\
  \bottomrule
  \end{tabular}
\end{table*}

\paragraph{Reading the tables.} Freq-Prune is the strongest prior baseline, but
\LorE{} preserves \emph{knowledge and reasoning} accuracy (MMLU, ARC-C, HellaSwag)
markedly better: it leads across the board on Qwen3, and on both models holds
MMLU/ARC well above the pruning and merging baselines, whereas D$^2$-MoE degrades
sharply at this expert count. On Gemma, Freq-Prune retains lower PPL and higher
commonsense (WinoGrande/PIQA), so \LorE{}'s advantage is in \emph{preserved
capability} rather than a clean sweep of every metric. As a side observation
consistent with the near-orthogonality analysis (Sec.~\ref{sec:analysis}),
\LorE{}'s margin over single-root decomposition (D$^2$-MoE) widens as the expert
count $E$ grows---the regime where one shared component is least able to fit the
experts.

\paragraph{Which variant to use.} The routing-aware variants win: the soft
cross-domain consensus $M_{\text{soft}}$ is strongest (best MMLU on both models
at $\sim$40--50\% compression), with Coact-NPMI close behind, because clustering
by co-activation groups experts that are functionally related rather than merely
close in weight space. Protecting the highest-firing experts also helps sharply
(WS-Frob MMLU $42.4\to53.7$ on Qwen3 with protK8), since those load-bearing
specialists are the costliest to approximate.

\subsection{Compression--quality frontier}
\begin{figure}[t]
\centering
\includegraphics[width=\colfigwidth]{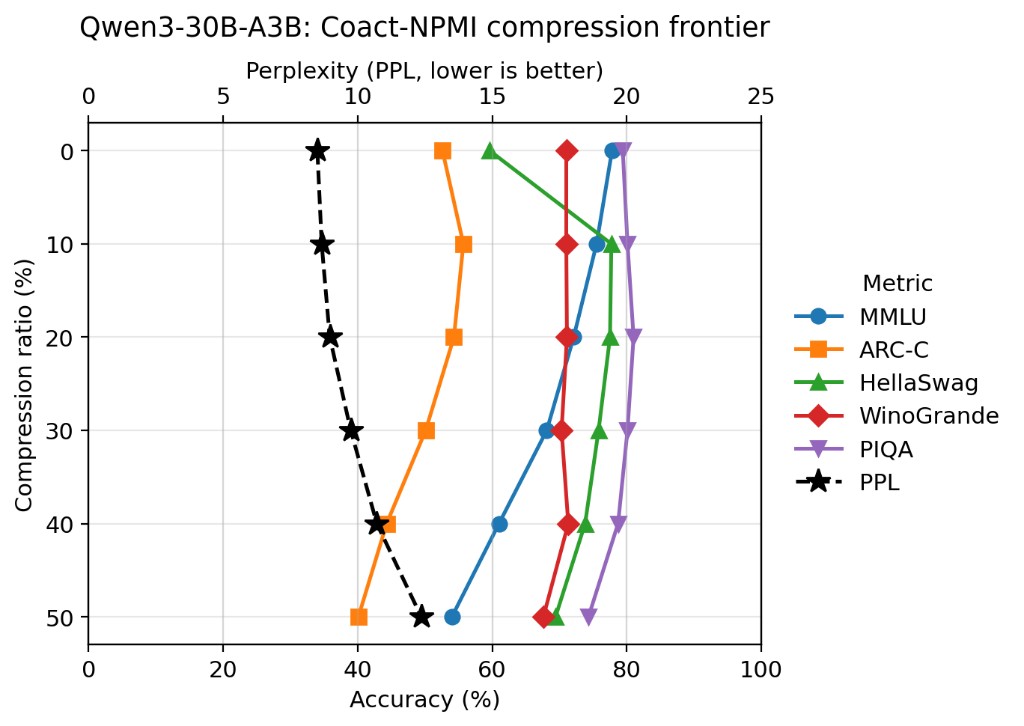}
\caption{Compression--quality frontier for \LorE{} (Coact-NPMI) on
  Qwen3-30B-A3B. As the expert compression ratio increases (top to bottom),
  downstream accuracy (MMLU, ARC-C, HellaSwag, WinoGrande, PIQA; bottom axis)
  stays near its uncompressed value through moderate compression and falls off
  mainly near $50\%$, while perplexity (dashed, top axis) rises gradually.}
\label{fig:frontier}
\end{figure}

The tables fix a single operating point, but the compression ratio is a knob we
can turn. In Figure~\ref{fig:frontier} we sweep the Coact-NPMI variant on
Qwen3-30B-A3B from $0$ to about $50\%$ expert compression and track perplexity
alongside the five downstream tasks. Accuracy changes little through low and
moderate compression, and the clearer drops set in only as the ratio nears
$50\%$; perplexity climbs slowly over the same range. Quality therefore falls
off smoothly rather than all at once, and the $50\%$ end of the sweep reproduces
the Coact-NPMI row of Table~\ref{tab:qwen3}. In practice the ratio can then be
tuned to a quality target rather than fixed in advance.

\subsection{Calibration cost}
Calibration time is dominated by fitting the low-rank factors, and this admits a
large, \emph{analytically guaranteed} speedup. The reconstruction fine-tuning of
Sec.~\ref{sec:finetuning} reaches the factors through $\sim$500 iterative gradient
steps over cached activations. That objective, however, is an
\emph{activation-weighted low-rank regression}, which has a \emph{closed-form}
optimum: whitening by the activation second moment and taking a single
(generalized) SVD of the residual recovers the optimal rank-$r$ factors directly.
Replacing the $\sim$500 optimization passes with one whitened SVD collapses the
factor-fitting stage from $O(T)$ passes to $O(1)$---roughly \emph{two orders of
magnitude} ($\sim$$100\times$) less compute for that stage---while attaining the
\emph{same} rank-$r$ optimum, so accuracy is unchanged. Because factor fitting
dominates the pipeline, this turns whole-model calibration into a short offline
step rather than a bottleneck. We report this speedup as a relative multiplier;
absolute wall-clock times are omitted.

%% file: sections/07_discussion.tex
\section{Discussion}
\label{sec:discussion}

\paragraph{Why \LorE{} works.} The gains trace to one design choice grounded in
Sec.~\ref{sec:analysis}: \emph{multiple local anchors} (per-cluster dominants)
instead of one global root, so each anchor is close to its members and the method
scales with $E$ where single-root decomposition degrades. This preserves all
experts and the original router.

\paragraph{Hessian-weighted fitting (ongoing work).} Sec.~\ref{sec:results}
introduces the activation-weighted closed-form solve as a way to \emph{replace}
the iterative reconstruction pass at equal quality. We note here that the same
device is also a quality improvement in its own right over the plain-SVD default,
and that it generalizes beyond the corrections.

The default fit minimizes weight-space error $\lVert W - \widehat{W}\rVert_F$,
which is not the quantity a layer emits. The error reaching the next layer is
$\lVert X(W - \widehat{W})^\top\rVert_F$ for the calibration activations $X$ the
router actually sends to that expert---a per-expert quantity in an MoE, since
each expert sees only its own slice of the tokens. With
$G_e = X_e^\top X_e = L_eL_e^\top$, minimizing the emitted error is again a
truncation problem---of $WL_e$ rather than $W$---so it costs the same rank, the
same storage $r(I+H)$, and the same two matmuls at inference. In preliminary runs
on Qwen3-30B-A3B this improves perplexity at matched storage for every variant we
have tried. One MoE-specific caveat: $G_e$ has $H^2$ entries estimated from the
tokens a single expert receives, so it needs substantially more calibration data
than a diagonal or norm-based statistic, and experts that see fewer tokens than
$H$ rely on damping.

The same reweighting extends to the shared components themselves, and there it
also removes a structural weakness. For a shared component $C$ serving expert set
$S$ with corrections $R_e$, the optimal Hessian-weighted choice is closed-form,
\[
  C^\star = \Big[\textstyle\sum_{e\in S}(W_e - R_e)\,G_e\Big]
            \Big[\textstyle\sum_{e\in S} G_e\Big]^{-1},
\]
which is a traffic-weighted combination of the cluster rather than a member of
it. Because $C=0$ lies in the feasible set, a solved anchor can never be worse
than using no anchor at all---a guarantee that selecting an existing expert as
the dominant does not provide, and one that matters precisely when experts are
near-orthogonal and the residual against another expert is larger than the
expert itself. Note that a solved anchor is no longer one of the experts, so the
former dominant also needs a correction; at $k{=}64$ this is roughly a $10\%$
increase in the expert budget, which must be returned by a small reduction in
$r$. We are pursuing the full treatment, including its interaction with the
\BTE{} tree, where the component values change but the sharing structure---and
hence the amortization factor---does not.

\paragraph{Limitations.}
\begin{enumerate}
  \item \textbf{Calibration cost.} Calibration is dominated by the per-member SVD
        and the short reconstruction pass; a closed-form activation-weighted solve
        can replace the SVD (Sec.~\ref{sec:results}, Gate H) but is not yet the default.
  \item \textbf{Inference amortization is bounded.} The \BTE{} amortization
        benefit is routing-dependent and partly shared with any
        shared-component method; we report only fairly-baselined numbers
        (Sec.~\ref{sec:btexperts}) and make no strawman speedup claims.
\end{enumerate}

%% file: sections/08_conclusion.tex
\section{Conclusion}
\label{sec:conclusion}

We presented \textbf{\LorE{}}, a router-preserving method for compressing
Mixture-of-Experts weight matrices. Motivated by a mechanistic analysis of the
``orthogonal experts'' problem---experts are near-orthogonal in weight space yet
organize into functional co-activation communities---\LorE{} keeps a full
dominant per cluster and represents other experts as low-rank corrections to that
local dominant, retaining all experts and the original router with no rerouting or
router retraining. A lightweight reconstruction fine-tuning refines the factors,
and \BTE{} organizes dominants/corrections for inference-time amortization.

Across Qwen3-30B-A3B and Gemma-4-26B-A4B at $\sim$50\% expert compression, \LorE{}
preserves downstream capability---particularly knowledge- and reasoning-heavy
accuracy (MMLU, ARC-C, HellaSwag)---better than pruning, merging, and
single-shared-component decomposition (D$^2$-MoE), and its advantage over the
latter grows with the expert count.

\paragraph{Future work.} Higher-compression and full-model operating points;
generation-benchmark evaluation; faster closed-form factor fitting at scale; and
using the \BTE{} structure for upcycling / MoE construction from dense
checkpoints.

%% file: appendices/C_ablations.tex
\section{Ablations}
\label{sec:ablations}

\paragraph{Tree vs.\ flat (quality).} A recurring question is whether organizing
the low-rank corrections as a hierarchy (\BTE{}) helps compression \emph{quality}
relative to flat per-cluster decomposition. Table~\ref{tab:treeflat} answers it in
a controlled study on Mixtral-8$\times$7B ($E{=}8$) at matched compression: the two
are indistinguishable within noise, both zero-shot and after reconstruction
fine-tuning. This confirms that \BTE{} is a quality-neutral \emph{inference-time}
organization (Sec.~\ref{sec:btexperts}), not a compression-quality mechanism---its
value is amortization, not accuracy.

\begin{table}[h]
\centering
\caption{Tree (\BTE{}) vs.\ flat per-cluster decomposition on Mixtral-8$\times$7B
  ($E{=}8$; layer~16, rank~32, 3 seeds), at matched compression. $\Delta$PPL is the
  increase over the uncompressed baseline (lower is better); the hierarchy matches
  flat decomposition within noise.}
\label{tab:treeflat}
\small
\setlength{\tabcolsep}{4pt}
\resizebox{\columnwidth}{!}{%
\begin{tabular}{l r r r}
\toprule
Method & comp & $\Delta$PPL (zero-shot) & $\Delta$PPL (+recon-FT) \\
\midrule
\BTE{} (tree)                   & $7.5\times$ & $0.112{\pm}0.012$ & $0.099{\pm}0.011$ \\
D\textsuperscript{2}-MoE (flat) & $7.4\times$ & $0.105{\pm}0.011$ & $0.098{\pm}0.010$ \\
\bottomrule
\end{tabular}%
}
\end{table}

%% file: appendices/B_coactivation.tex
\section{Additional Co-Activation Views}
\label{sec:coact_appendix}

\begin{figure}[t]
\centering
\includegraphics[width=\colfigwidth]{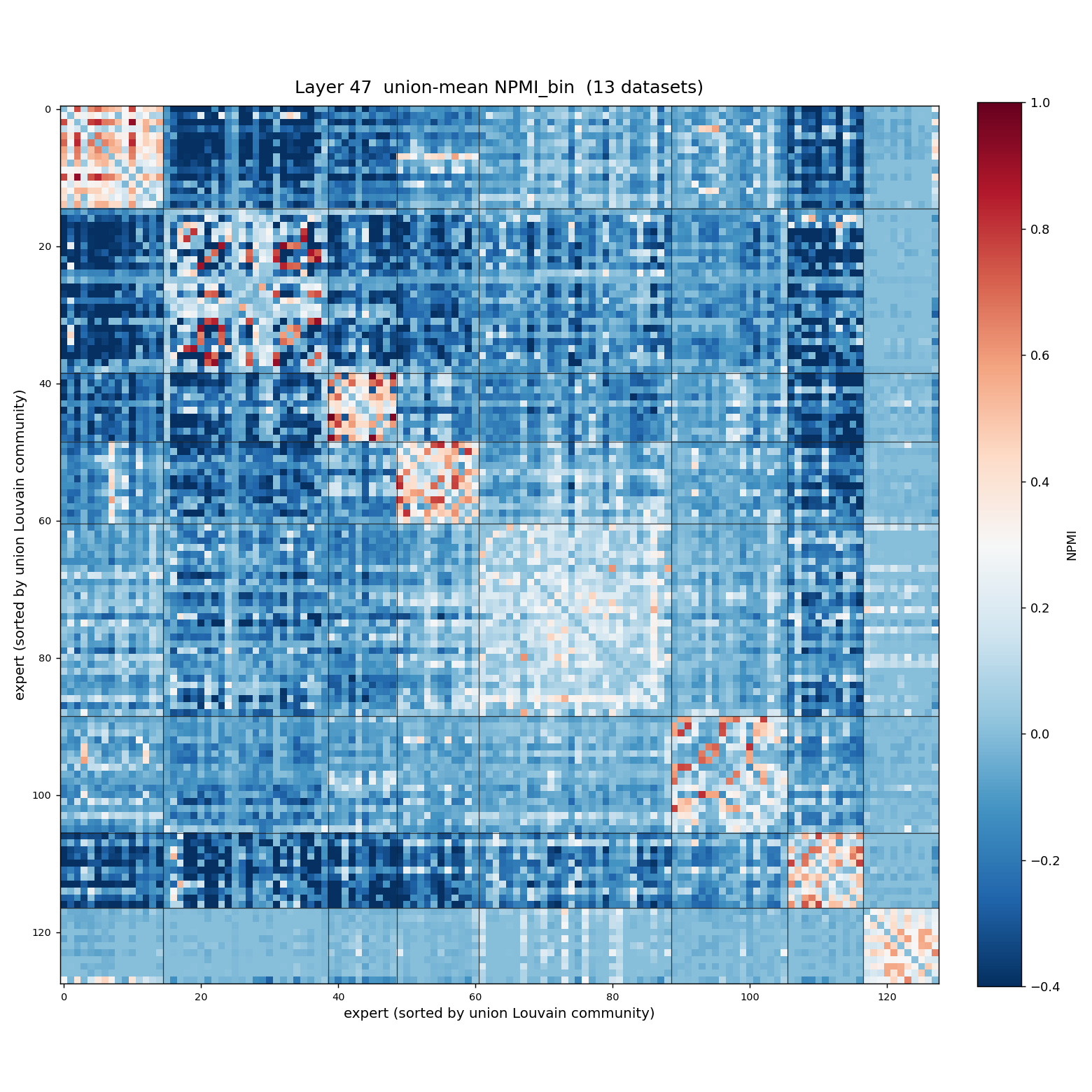}
\caption{Union-mean NPMI co-activation matrix at Qwen3-30B-A3B layer~47 (over 13
  datasets), experts reordered by community: co-firing experts form clear diagonal
  blocks even though their weights are near-orthogonal.}
\label{fig:coact_heat}
\end{figure}

\begin{figure}[t]
\centering
\includegraphics[width=\colfigwidth]{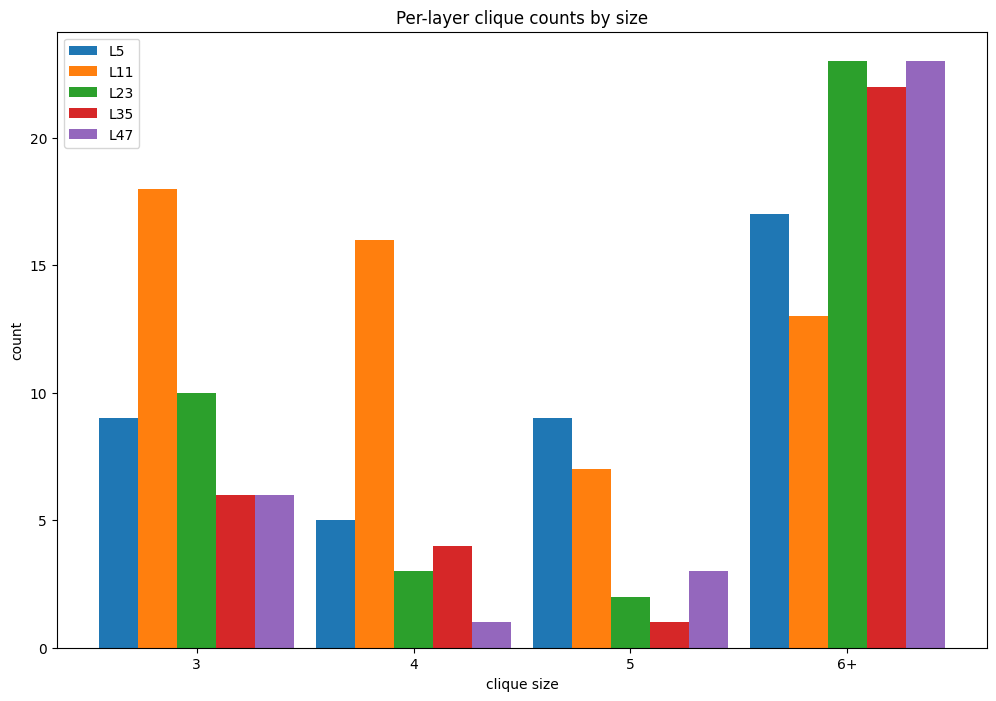}
\caption{Per-layer counts of dense expert cliques by size (Qwen3-30B-A3B). Both the
  number and the size of co-activation cliques vary with depth, so a uniform
  per-layer compression budget is suboptimal.}
\label{fig:coact_cliques}
\end{figure}

The backbone graph in the main text (Fig.~\ref{fig:npmi}) summarizes the
co-activation structure at layer~23. Figure~\ref{fig:coact_heat} shows the same
kind of structure as a raw NPMI heatmap at a late layer, and
Fig.~\ref{fig:coact_cliques} reports how the dense co-activation cliques shift with
network depth.

